\documentclass[conference]{IEEEtran}
\ifCLASSINFOpdf
\else
\fi
\usepackage{amsfonts}

\ifCLASSOPTIONcompsoc
    \usepackage[caption=false, font=normalsize, labelfont=sf, textfont=sf]{subfig}
\else
\usepackage[caption=false, font=footnotesize]{subfig}
\fi
\usepackage{mathrsfs}
\usepackage[ruled,linesnumbered]{algorithm2e}
\usepackage{graphicx}

\usepackage{amsmath}
\usepackage{xcolor}
\newcommand{\R}{\mathbb R}
\newcommand{\mG}{\mathcal G}
\newcommand{\mB}{\mathcal B}
\newcommand{\bw}{\textbf w}
\newcommand{\bS}{\textbf S}
\newcommand{\bs}{\textbf s}
\newcommand{\bL}{\textbf L}
\newcommand{\bW}{\textbf W}

\newcommand{\argmin}[1]{\underset{#1}{\text{argmin}}}
\newcommand{\prox}{\textbf{prox}}
\newcommand{\proj}{\textbf{proj}}

\SetKwInput{KwInit}{Initialize}
\SetKwInput{KwMain}{Main Loop}

\begin{document}
%
\title{Latent Group Structured  Multi-task Learning}

\author{\IEEEauthorblockN{Xiangyu Niu}
\IEEEauthorblockA{Department of Electrical Engineering \\and Computer Science\\
University of Tennessee, Knoxville\\
Knoxville, Tennessee 37996\\
Email: xniu@vols.utk.edu}
\and
\IEEEauthorblockN{Yifan Sun}
\IEEEauthorblockA{Department of Computer Science\\
University of British Columbia\\
Vancouver, B.C. V6T 1Z4 \\
Canada\\
Email: ysun13@cs.ubc.ca}
\and
\IEEEauthorblockN{Jinyuan Sun}
\IEEEauthorblockA{Department of Electrical Engineering \\and Computer Science\\
	University of Tennessee, Knoxville\\
	Knoxville, Tennessee 37996\\
	Email: jysun@utk.edu}}


%


\maketitle

\begin{abstract}
In multi-task learning (MTL), we improve the performance of key machine learning algorithms by training various tasks jointly.
When the number of tasks is large, modeling task structure can further refine the task relationship model. 
For example, often tasks can be grouped based on metadata, or via simple preprocessing steps like K-means. 
In this paper, we present our group structured latent-space multi-task learning model, which encourages group structured tasks defined by prior information.
We use an alternating minimization method to learn the model parameters.
Experiments are conducted on both synthetic and real-world datasets, showing competitive performance over single-task learning (where each group is trained separately) and other MTL baselines.
\end{abstract}


%
\IEEEpeerreviewmaketitle

\section{Introduction}
Multi-task learning (MTL) \cite{caruana1998multitask,argyriou2007multi} seeks to improve the performance of a specific task by sharing information across multiple related tasks. Specifically, this is done by simultaneously training many tasks, and promoting relatedness across each task's feature weights.
MTL continues to be a promising tool in many applications including medicine \cite{wang2018obesity,zhang2018identifying}, imaging \cite{gan2018multiple}, and transportation \cite{wang2018joint}, and has recently become repopularized in the field of deep learning (e.g. \cite{zhang2014facial,wu2015deep, collobert2008unified, caruana1993multitask}).

A common approach to  multi-task learning \cite{argyriou2007multi,evgeniou2005learning} is to
provide task-specific weights on features. 
There are two main approaches toward enforcing task relatedness. One is by promoting similarity between task weights, either through regularization  in convex models \cite{argyriou2008convex,kang2011learning,evgeniou2004regularized} or  structured priors in statistical inference \cite{yu2005learning,yang2013multi,lee2007learning}.
The other is through enforcing a \emph{low-dimensional latent space } representation of the task weights \cite{zhou2011clustered,kumar2012learning,ruvolo2014online,ando2005framework, daume2009bayesian}. 

Where the main strength of multi-task learning is shared learning across tasks, a main weakness is contamination from unrelated tasks. 
Many MTL models have been proposed to combat this contamination.
Weighted task relatedness can be imposed via the Gram matrix of the features \cite{wang2009semi} or a kernel matrix \cite{evgeniou2005learning}, or probabilistically using a joint probability distribution with a  given covariance matrix.
Pairwise relativeness can also be learned by optimizing for  a sparse covariance or inverse covariance matrix expressing the relatedness between weights on different tasks; this can be done either by alternating minimization \cite{zhang2010convex,gonccalves2016multi,jacob2009clustered} or variational inference\cite{yang2013multi}.
To encode specifically for group or cluster structure, one approach is to compose a combinatorial problem, in which group identity is represented by an integer \cite{kang2011learning,argyriou2008algorithm}.
Another approach is to provide multiple sets of weights for each task, regularized separately to encode for different hierarchies of relatedness \cite{han2015learning}.

For latent space models,   task relatedness and grouping can be imposed  more simply, without many added variables or discrete optimization. Concretely,
each task-specific feature vector  can be modeled as $x^t \approx Ls^t$, where the columns of $L$ encode the latent space and a sparse vectors $s^t$ decide how the latent vectors are shared across tasks. 
In this context, \cite{ruvolo2014online} uses dictionary learning to  learn a low rank $L$ and sparse $s^t$'s for each task; this representation is then applied to online learning tasks. 
Similarly,
\cite{kumar2012learning} uses the same model, but learns $L$ and $s^t$ through minimizing the task loss function.
In both cases, the support of $s^t$ for different $t$'s may overlap, enabling this model to capture overlapping group structure in a completely unsupervised way.



In this paper, we take a step back and decompose this problem into two steps: first, learning task relatedness structure, and then, performing MTL. The main motivation for this two-step approach is that oftentimes, task-relatedness already given in the metadata. 
For  example, 
in the task of interpreting geological photos, it may be helpful to use terrain labels (forest, desert, ocean) to help group the types of images.
Similarly, 
in regressing school test scores, where each task corresponds to a school, a task group may be a specific district, or designation (private/public/parochial), or a cluster discovered from ethnic or socioeconomic makeup--in this way, the groups are intentionally interpretable. 
Once the (possibly overlapping) groups are identified, we solve the latent-space model with an overlapping group norm regularization on the variables $s^t$. 
Although the first step can be made significantly more sophisticated, we find this simple approach can already give superior performance on several benchmark tasks.


\section{Notation} For a vector $x\in \R^n$, we denote $x_i$ as the $i$th element of $x$, and for some group $G\subseteq \{1,\hdots, n\}$, we denote $x_G = \{x_i\}_{i\in G}$ the subvector of $x$ indexed by $G$. We denote the Euclidean projection of a vector $x$ on a set $\mathcal S$ as $\proj_{\mathcal S}(x)$. For a vector or matrix, we use $x'$ to represent the transpose of $x$. 
\section{Multi-task Learning}

\subsection{Linear Multi-Task Learning}

Consider  $T$ tasks,  each with $n_t,\; t = 1,2,...,T$ labeled training samples $\{ x_i^t\in \R^d, y_i^t\in \R \} _ {i= 1,2,...,n_t}$. 
Following \cite{argyriou2007multi}, 
we minimize the function
\begin{equation}
	\min_{\{\bw^t\}_t} \left \{ \sum_{t=1}^{m} \left( \sum_{i=1}^{n_t} \ell ((\bw^t)' x_i^t, y_i^t) \right) + \mathcal{R} (\bw^{(1)},\hdots,\bw^t)  \right \}
\end{equation}
over task-differentiated weights on each feature $w^t$, differentiated per task. 
The  loss function $\ell$ is any smooth convex loss function, such as squared loss for regression or logistic loss for classification. 
The regularization $\mathcal R$ enforces task relatedness.



\subsection{Latent Subspace Multi-task Learning}



In latent subspace MTL \cite{ruvolo2014online,kumar2012learning}, task relatedness is expressed through a common low-dimensional subspace $\bw^t = \bL\bs^t$, where $\bL \in \R^{d\times k}$ is common to all tasks, and a sparse $\bs^t$ weights the component of each subspace to each task.
In \cite{ruvolo2014online,kumar2012learning}, parsimony is enforced via the regularization:
\[
\mathcal R(\bL,\{\bs^t\}_t) = \mu \|\bL\|_F^2 + \lambda\sum_{t=1}^T \|\bs^t\|_1.
\]
Here, $\ \| \textbf{L}\|_{F}^2 = \sum_{ij}L_{ij}^2$ is the square of the Frobenius norm of $L$ and is frequently used to promote low-rank (since $\|L\|_F^2 = \text{tr}(L'L)$) and $\|\bs^t\|_1 = \sum_i |\bs^t_i|$ promotes sparsity on each  $\bs^t$. In  words, the weights $\bw^t = \bL\bs^t$ for different $t=t_1$ and $t=t_2$ are related through latent space $\bL_k$ only if $\bs^{t_1}_k$ and $\bs^{t_2}_k$ are nonzero. 
 
\subsection{Group Structured Latent Subspace Multi-task Learning}

We now describe our group-regularized latent space MTL model. As before, we assume task parameters $\textbf{w}_t$ within a group lie in a low-dimensional subspace, and the penalty function $\mathcal R$  promotes group structure.
We first assign the set of tasks into $g$ groups $\mG = \{G_1,\hdots, G_g\}$. The assignment may not be unique; i.e. we allow for groups with overlaps.
The weight parameter of every task is a linear combination of $k < T$ latent tasks. 
Mathematically, stacking $\bW = [\bw^{1},\hdots, \bw^t]$ and $\bS = [\bs^{1},\hdots, \bs^t]$, we can describe $\bW = \bL\bS$ and 
\[
\mathcal R(L, \{\bs^t\}_t) = \mu\|\bL\|_F^2 + \lambda \|\bS\|_{\mG,1}, \quad 
\|\bS\|_{\mG,1} = \sum_{i=1}^d  \|\bS_{i,:}\|_\mG
\]
where $\bS_{i,:}$ is the $i$th row of $\bS$.
and 
\[
\|x\|_\mG = \min_{w_1,\hdots,w_g}\left\{\sum_{G\in \mG} \|w_G\|_2 : x = \sum_{G\in \mG} w_G \right\}
\]
is the  \emph{group norm} proposed in  \cite{jacob2009group}.
The overall optimization function can be expressed as

\begin{equation} \label{loss_fun}
\min_{\textbf{L}, \textbf{S}} \sum_{t=1}^T \sum_{i=1}^{N_t} \ell (y_i^t, (\textbf{s}^t)' \textbf{L}' x_i^t) + \mu \left \| \textbf{S} \right \|_{\mG,1} +  \lambda  \| \textbf{L}  \|_{F}^2.
\end{equation}


We evaluate our model for two tasks: 
\begin{enumerate}
\item linear regression, where $\ell(y_i^t, \bs_t'\bL'x_i^t) = (\bs_t'\bL'x_i^t-y_i^t)^2$, and 
\item logistic regression, where $\ell(y_i^t, \bs_t'\bL'x_i^t) = \log (1+\exp(-y_i^t \bs_t'\bL'x_i^t))$.
\end{enumerate}

\subsection{Generalization to basis functions}
The success of each of these models is based on the assumption that a linear representation of the feature space is sufficient for modeling, e.g. 
\begin{equation}
y_i^t \approx (\bw^t)' x_i^t
\label{e-linearmodel}
\end{equation}
Note that this representation can easily be made nonlinear through a set of nonlinear  functions $\phi_1,\hdots, \phi_d$, with
\begin{equation}
y_i^t \approx \sum_{j=1}^d (\bw^t)' \phi_j(x_i^t)
\label{e-basismodel}
\end{equation}
which, for known $\phi_j$, does not increase the numerical difficulty. Therefore, although the model \eqref{e-basismodel} is more general, for clarity we restrict our attention to the model \eqref{e-linearmodel}.


\section{Optimization Procedure}
Although \eqref{loss_fun} is not convex, it is biconvex in $\bL$ and $\bS$; we therefore solve \eqref{loss_fun} using an alternating minimization strategy:
\begin{eqnarray}
\textbf{L}^+ &=& \argmin{\bL} \sum_{t=1}^T \sum_i \ell (y_i^t, (\bs^t)' \bL' x_i^t) +  \lambda  \| \textbf{L}  \|_{F}^2\label{opti_l}\\
\textbf{S}^+ &=& \argmin{\bS} \sum_{t=1}^T \sum_i \ell (y_i^t, (\bs^t)'(\bL^+)' x_i^t) + \mu  \| \textbf{S}  \|_{G}\label{opti_s}
\end{eqnarray}
where $\bS^+,\bL^+$ are the new iterates.
The optimization function in \eqref{opti_l} is over a smooth, strongly convex function in $\bL$, and can be efficiently minimized either via accelerated gradient descent   directly via backsolve if $\ell(y,x) = \|x-y\|_2^2$ (e.g. linear regression). 

The optimization for $\bS$ in \eqref{opti_s} is less straightforward, since the group norm is nonsmooth; this is done  via the fast iterative shrinkage-thresholding algorithm (FISTA), which to minimize $f(\bS)+g(\bS)$
where $f(\bS) = \sum_t \sum_i \ell(y_i^t, \bs_t'\bL'x_i^t)$ is convex and differentiable, and $g(\bS) = \|\bS\|_{1,\mathcal G}$ is nonsmooth. The FISTA iterates  for a step size $t > 0$ are then
\[
\bS_i^+ = \prox_{tg}(\bS_i-t\nabla f(\bS_i)) 
\]
for each row $i = 1,\hdots, k$ of $\bS$. 
The \emph{proximal operator}\cite{moreau1965proximite} is defined as
\[
\prox_{tg}(x) =  \argmin{u} g(u) + \frac{1}{2t}\|u-x\|_2^2.
\]

For $g(x) = \|x\|_{\mG}$, we compute $\prox_{tg}(x)$ as described in \cite{villa2014proximal}. Define the sets $\mB_k$ where $x\in \mB_k \iff \|x_{G_k}\|_2\leq 1$, for $k = 1,\hdots, g$. Define $\mB = \mB_1 \cap \mB_2 \cap \cdots \cap \mB_g$. That is, 
\[
x\in t \mathcal B \iff \|x_{G_k}\|_2 \leq t, \; k = 1,\hdots, g.
\]
By Fenchel duality, 
\[
\prox_{tg}(x) = x - \proj_{ t \mathcal B}(x).
\]
Note that if the groups are not overlapping, then this projection decomposes to $g$ smaller projections:
\[
z_{G_k} := \proj_{t\mB_k}(x_{G_k}) = \frac{t}{\|x_{G_k}\|_2}x_{G_k}
\]
and can be computed in one step.
When the groups overlap, we adopt the simple cyclic projection algorithm proposed in \cite{villa2014proximal,bauschke1996approximation}. These steps are outlined precisely in Alg. \ref{alg-prox}.

The algorithm listing in Algorithm~\ref{alg} outlines the alternating minimization steps for solving Equation~\ref{loss_fun}. We adopt the method from \cite{kumar2012learning} to initialize $\textbf{L}$. The individual task weight matrix $\textbf{w}_t$ are first learned independently using their own data $\textbf{Z}_t$, and stacked  as the columns in $\textbf{W}$. The matrix $\textbf{L}$ is then initialized with the top-k singular vectors of $\textbf{W}$. 
The main algorithm then alternately minimizes for $\bL$ and $\bS$, and  is terminated when there is small change of either $\textbf{L}$ or $\textbf{S}$ between  consecutive iterations

\begin{algorithm}[!t] \label{alg}
	\KwData{$(x^t_i,y^t_i)$, $i=1,\ldots,n_t$ samples, $t = 1,\ldots,T$ tasks. 
		$k$: Number of latent tasks.}
	\textbf{Initialization}
    Learn individually for each task
    \[
    \textbf w_t = \argmin{w} \sum_{i=1}^{n_t} \ell(y_i^t ; \textbf w_t'x_i^t), \quad \bW = [\textbf w_1,\ldots,\textbf w_T].
    \]\\
	Compute top-$k$ SVD: $\textbf{W} = \textbf{U}_k \mathbf{\Sigma}_k \textbf{V}_k'$. \\
	Initialize $\bL = \textbf U_k$\\
	\While{not converged}{
		Fix $\textbf{L}$ and solve for $\bS$ via Alg. \ref{alg-prox}\;
		Fix $\textbf{S}$ and solve for $\bL$ 
        \[
        \textbf{L} = \argmin{\textbf{S}} \sum_{t=1}^T \sum_i^{n_t} l (y_t^i, \textbf{r}'_t \textbf{L}' x_t^i) +  \lambda \left \| \textbf{L} \right \|_{F}^2
        \]
        via gradient descent or direct method\;
	}
	\KwResult{Task predictor matrices $\bL$ and $\bS$.}
	\caption{Group structured subspace multi-task learning}
\end{algorithm}

\begin{algorithm}[!t] \label{alg-prox}
	\KwData{Current $\bS$, $\bL$, step size $t >0$}

	\While{not converged}{
		Take a gradient step $\tilde \bS = \bS - t \nabla f(\bS)$\\
        \For{  $i=1,..,d$}{
        \% Compute the projection of row $\tilde\bS_{i,:}$on $\mB$\\
        \If{groups are separable}{
        $(\bS_i)_{G_{j'}} :=  \frac{t}{\| (\tilde \bS_i)_{G_{j'}}\|}  (\tilde \bS_i)_{G_{j'}}$
        }
        \Else{
        $z^{(0)} = 0$\\
        \For{$j= 1 \ldots$ until convergence}{
            $j' = j \text{ mod } g$\\
            $z^{(j+1)}_{G_{j'}} := \frac{1}{j+1} (\tilde \bS_i)_{G_{j'}} + \frac{j}{j+1} \frac{t}{\| (z^{(j)}_{G_{j'}})\|}  (z^{(j)}_{G_{j'}})$.
            
        }
        Update $\bS_i = z^{(j)}$\\
        }
        
	}
    }
	\KwResult{$\textbf{S} = [\bS_1',\bS_2',\ldots,\bS_k']$.}
	\caption{Update of $\bS$}
\end{algorithm}

\section{Experiments}

In this section we provide experimental results to show the effectiveness of the proposed formulation for both regression and classification problems.
We compare against three baselines:
\begin{itemize}
\item \textbf{Single task learning (STL)}: Each task is learned separately, through logistic or ridge regression.
\item \textbf{MTL-FEAT} \cite{argyriou2008convex}: A latent space MTL model that regularizes $\bS$ via the 2,1 norm, e.g. solves
\[
\min_{\bS,\bL} \sum_{t=1}^T\left(\sum_{i=1}^{n_t}\ell(y_i^t,\bs_i'\bL'x_i^t) + \mu \|s_t\|_1^2\right)
\]
\item \textbf{GO-MTL} \cite{kumar2012learning}: 
A latent space model that regularizes $\bL$ for low rank and $\bS$ for sparsity, e.g. minimizing
\[
\min_{\bS,\bL} \sum_{t=1}^T\sum_{i=1}^{n_t}\ell(y_i^t,\bs_i'\bL'x_i^t) + \mu \|\bS\|_1 + \lambda \|\bL\|_F
\]
\item \textbf{GS-MTL} Our model.
\end{itemize}

Below, we test three MTL models on synthetic and  real-world datasets. For each dataset, we apply a 60\% / 20\% / 20\%  train / validation / test split, with a grid search to determine the best $\lambda$ and $\mu$ over powers of 10. 

\subsection{Synthetic data}
We evaluate our model on two synthetic datasets: one we generate ourselves, and another baseline from a similar work, for comparison. In both, the task is regression.

\textbf{Synthetic 1} 
Consider $m$ the number of features and $\mG = \{G_1,G_2,\hdots,G_g\}$ the set of groups, where each $G_k \subset\{1,\hdots,m\}$. We uniformly sample cluster centers $\mu_1,\hdots, \mu_g$ where $\mu_k \in \mathcal U(0,1)^m$. Then, we generate $n$ datapoints $x\in \R^m$ 
such that $x_i\in \mathcal N(\mu_k,\sigma)$ where $i\in G_k$. If $i$ belongs to more than one group, then $k$ is picked randomly (uniformly) from the set $\{k : i\in G_k\}$. This is done independently for $i = 1,\hdots, m$. Specifically, $m = 20$,  $g = 3$, with 10 tasks and 20 samples per task.
The motivation behind this procedure is to model data vectors where each feature is drawn from a different clustering scheme--for example, a person's age, socioeconomic class, and geographic location produces different communities for which the person may belong, and each community characteristic plays a role in predicting the person's task performance.

\textbf{Synthetic 2 \cite{kumar2012learning}} We borrow the  synthetic dataset  from \footnote{https://github.com/wOOL/GO\_MTL},  to compare against other models. It consists of  $m=20$ dimensional feature vectors, 10 tasks, with 65 samples per tasks.



\subsection{Real datasets}
\begin{itemize}
\item Human Activity Recognition \footnote{https://archive.ics.uci.edu/ml/datasets/Human+Activity+Recognition+Using\\+Smartphones}: 
This dataset contains signal features from smartphone sensors held by test subjects performing various actions. The task is to classify the action based on the signal. In total there are 30 volunteers and 6 activities: walking, walking upstairs, walking downstairs, sitting, standing, and laying. Each datapoint contains 561 processed features from the raw signal. 
We pick the groups using a K-Means clustering of these feature vectors. 
We model each individual as a separate task and predict between sitting and other activities. 

\item Land Mine\footnote{http://www.ee.duke.edu/~lcarin/LandmineData.zip}: This dataset consists of 29 binary classification tasks. Each instance consists of a 9-dimensional feature vector extracted from radar images taken at various locations. Each task is to predict whether landmines are present in a field, where several images are taken.
The data is also labeled by terrain type: the first 15 fields are highly foliated (forests), while the last 14 are barren (desert). This lends to a natural group assignment.
\end{itemize}

The results are summarized in Table \ref{tab_result}. All MTL approaches have lower RMSE than the single task learning baseline, which confirms that sharing task information is crucial for optimal performance.  Moreover, the proposed method is able to outperform both MTL-FEAT and GO-MTL, suggesting successful incorporation of side information.

\begin{figure}[!t]
\centering
    \subfloat[Sparse pattern generated by GO\_MTL]{\label{fig:scen1}\includegraphics[scale=0.25, bb=0 0 480 480]{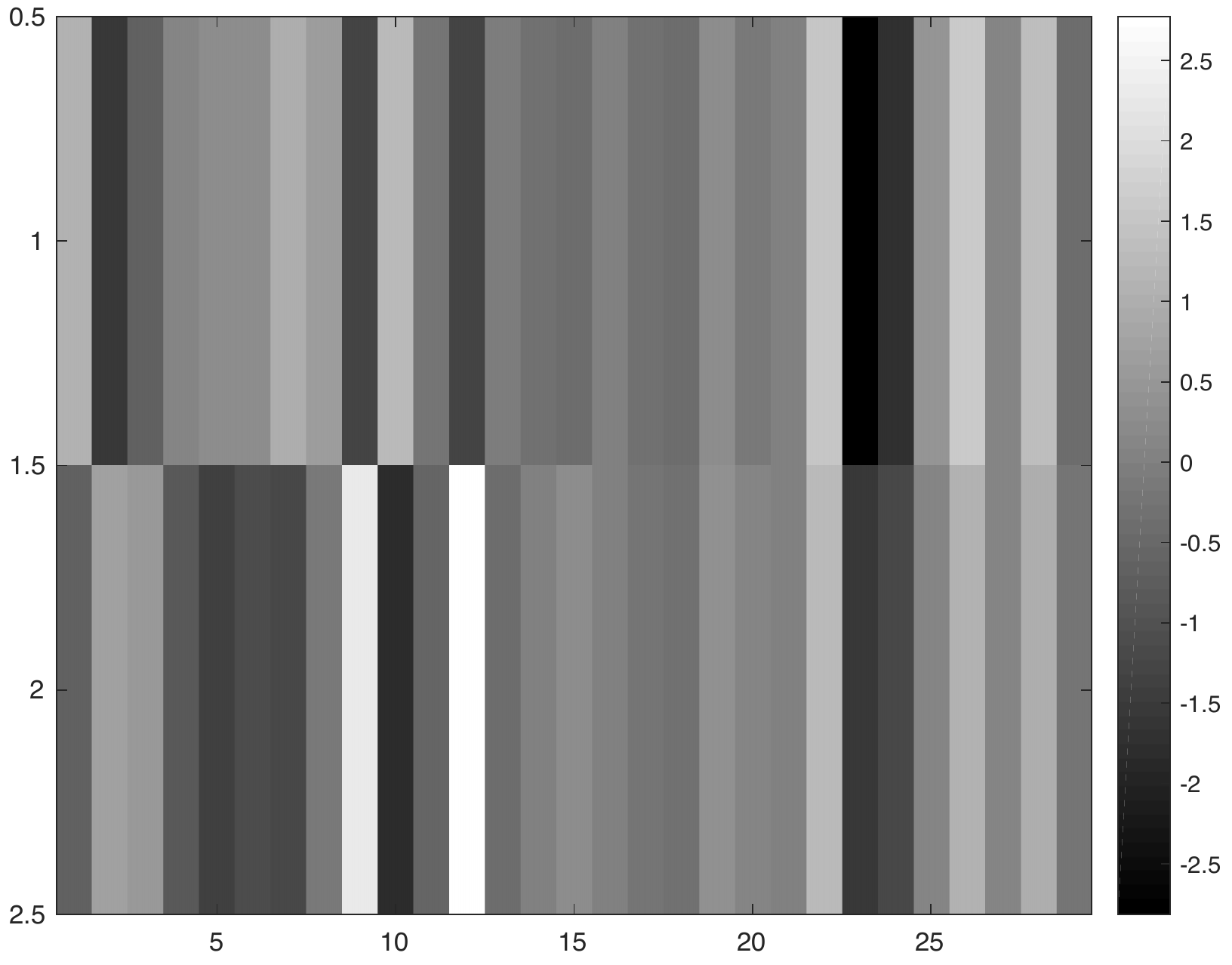}} 
    \subfloat[Sparse pattern generated by GS\_MTL]{\label{fig:scen2}\includegraphics[scale=0.25, bb=0 0 480 480]{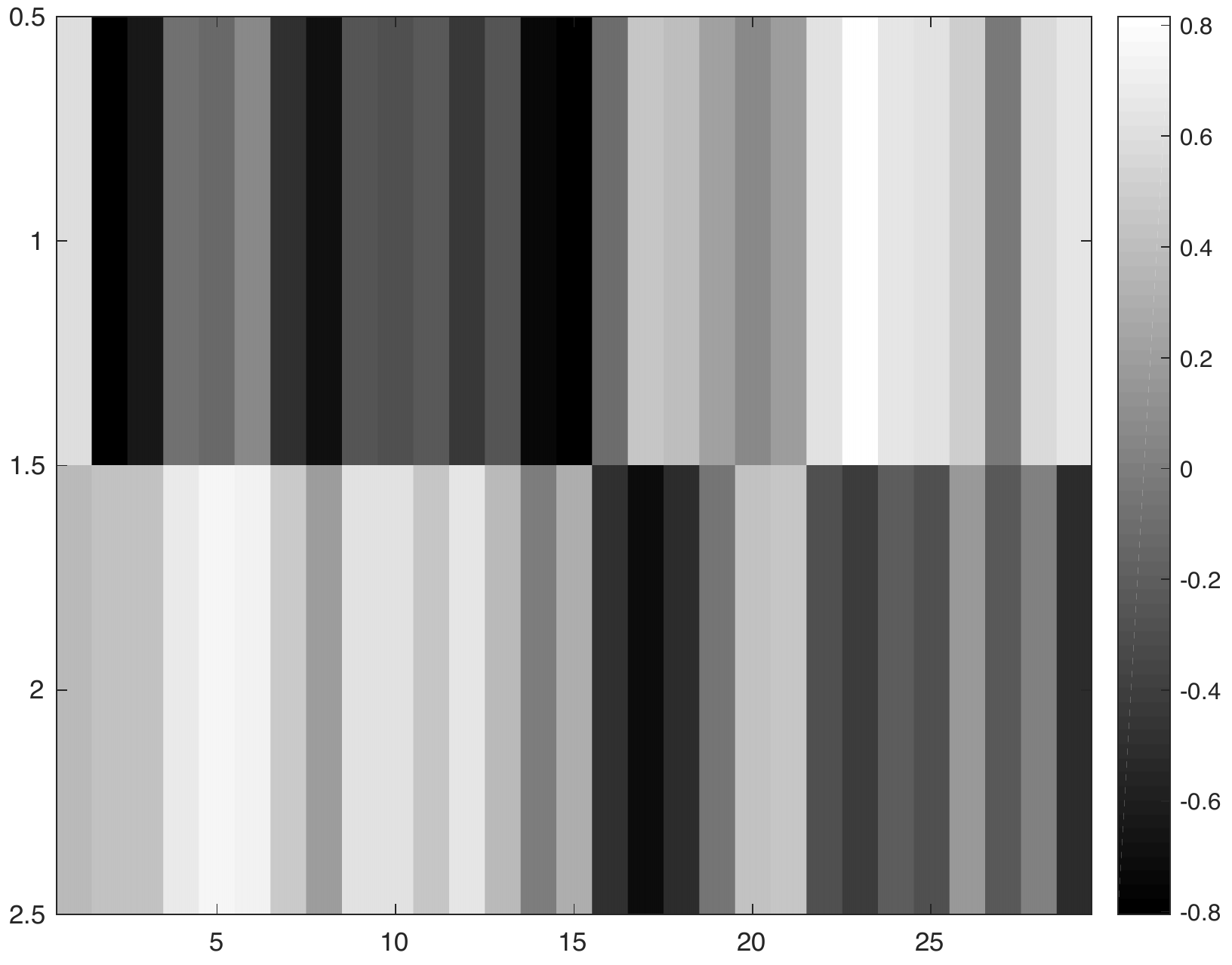}} 
\caption{Visualization of final $S$ matrix solved over the Landmine dataset. The 29 tasks are along the $x$-axis, and the 2 latent tasks are along the $y$ axis. 
Although the GO\_MTL model is designed to find group structure, it is difficult to discern. 
In our model, by specifically regularizing for the two groups, we see much clearer separation. However, the pattern is not completely block diagonal, showing some cross-group information sharing between tasks. 
}
\label{fig_sim}
\end{figure}

\begin{table}[!t]
\centering
\caption{Average prediction error for regression and classification datasets.}
\label{tab_result}
\begin{tabular}{|c|c|c|c|c|}
\hline
         & Synthetic1 & Synthetic2 & Landmine & Human Activity \\ \hline
STL      & 1.729    & 1.682     & 0.253        & 0.660    \\ \hline
MTL-FEAT & 1.553    & 1.099     & 0.292        & 0.641    \\ \hline
GO-MTL   & 1.314    & 0.430     & 0.240        & 0.580    \\ \hline
GS-MTL   & 1.253    & 0.385     & 0.2303        & 0.559     \\ \hline
\end{tabular}
\end{table}

\section{Conclusion}
In this paper, we proposed a novel framework for learning tasks' relationship in multi-task learning, where a prior group structure is either determined beforehand by problem metadata or inferred via an independent method such as K-means. We build upon models that enforce task interdependence through a latent subspace, regularized accordingly to capture  group structure.
We give algorithms for solving the resulting nonconvex problem for overlapping and non-overlapping group structure, and demonstrate its performance on simulated and real world datasets.
\end{document}